\documentclass{article}

\usepackage[utf8]{inputenc}

\PassOptionsToPackage{round}{natbib}
\usepackage[final]{tackling_climate_workshop_style}
\usepackage{amsmath,amsthm,amssymb}
\usepackage{mathtools}
\usepackage{dsfont}
\usepackage{graphicx}
\usepackage{wrapfig}
\usepackage{caption}
\usepackage{subcaption}
\usepackage{adjustbox}
\usepackage{todonotes}
\usepackage{xcolor}
\usepackage{url}

 \usepackage{enumitem}
 \setlist{topsep=0pt, leftmargin=*, itemsep=0pt}

\title{Land Use Prediction using Electro-Optical to SAR Few-Shot Transfer Learning }
\author{Marcel Hussing \hspace{2em} Karen Li \hspace{2em} Eric Eaton
\\
University of Pennsylvania
\\
\texttt{\{mhussing,\! karentli,\! eeaton\}@seas.upenn.edu}}

\date{}

\begin{document}

\maketitle

\begin{abstract}
Satellite image analysis has important implications for land use, urbanization, and ecosystem monitoring. Deep learning methods can facilitate the analysis of different satellite modalities, such as electro-optical (EO) and synthetic aperture radar (SAR) imagery, by supporting knowledge transfer between the modalities to compensate for individual shortcomings. Recent progress has shown how distributional alignment of neural network embeddings can produce powerful transfer learning models by employing a sliced Wasserstein distance (SWD) loss. We analyze how this method can be applied to Sentinel-1 and -2 satellite imagery and develop several extensions toward making it effective in practice. In an application to few-shot Local Climate Zone (LCZ) prediction, we show that these networks outperform multiple common baselines on datasets with a large number of classes. Further, we  provide evidence that instance normalization can significantly stabilize the training process and that explicitly shaping the embedding space using supervised contrastive learning can lead to improved performance.
\end{abstract}

\section{Introduction} 
The United Nations has estimated that a large proportion of the Sustainable Development Goal indicators can be measured through use of geospatial data~\citep{un2019geospatial}.  
As an example, the goal of land use and land cover mapping is to measure the health of populations, urban areas, and ecosystems over time. Since urban areas are responsible for approximately $70\%$ of the world’s energy-related CO$_2$ emissions, tracking the development of cities plays a crucial role in climate change mitigation and adaptation \citep{lucon2014buildings}. 
Interest in the topic has given rise to the collection of different forms of satellite data as a means to map out geospatial regions of the earth. One common form are hyperspectral electro-optical (EO) images, for which there exists an abundance of large labeled datasets. However, EO images cannot capture many factors of our earth's development; EO sensors are blocked by clouds, limited by the day-and-night cycle, and subject to distortion under various weather conditions. In situations that require imaging over extended time periods, researchers have started working with Synthetic Aperture Radar (SAR) data. SAR data is collected by a satellite sending out radar energy and recording the amount reflected back from the earth, making it insusceptible to many of the downsides of EO imaging. Unlike EO datasets, which can be easily interpreted and labeled through crowdsourcing, SAR data requires trained expertise for interpretation. 
As a result, labeled SAR datasets are both limited in quantity and costly to generate. 

One approach to extract information from geospatial data with few available labels is transfer machine learning. The idea is to pre-train a neural network on a source domain with plentiful data (i.e., EO) and then fine-tune on a target domain for which little data is available (SAR). In this paper, we focus on a recent method by~\citet{rostami2019transfer, rostami2019fewshot}, which uses sliced Wasserstein distance to align the distributions of the two modalities in a shared embedding space, and promote knowledge transfer between the classification of EO and SAR images. In doing so, one is not only able to use the few labeled SAR data, but also make use of any available unlabeled data. The original work is limited in that it only does analysis on binary classification using a relatively simple dataset---ships on uncluttered backgrounds. More realistic settings typically involve multiple classes and cluttered scenes, presenting a substantially harder challenge.  
Further, the original papers do not explore what properties of the embedding space might lead to better alignment, required for these real-world tasks.

To gain insight for how to apply EO-to-SAR transfer in practice, this paper studies these issues of multi-class classification and properties of the embedding space.  We performed the analysis using Sentinel-1 (EO) and Sentinel-2 (SAR) satellite data, and provide evidence that:
\vspace{-0.5em}
\begin{itemize}
    \setlength\itemsep{-2pt} 
    \item SWD embedding alignment for EO-to-SAR transfer can be scaled to the multi-class setting, 
    \item Applying instance normalization leads to more stable training and better performance, and
    \item Applying constrastive learning improves transfer performance of the SWD approach.
\end{itemize}

\section{Data \& Methods}

{\bf So2Sat Dataset}~~ 
Focusing on local climate zone classification, our analysis uses the So2Sat LCZ42 dataset ~\citep{zhu2020so2sat}, which consists of LCZ labels for approximately half a million Sentinel-1 (EO) and Sentinel-2 (SAR) satellite image patches over 42 cities across the world.  Each patch has a resolution of $32 \times 32$ pixels, and is labeled with one of 17 total classes (ten urban zone types, seven natural zone types). Each zone type is identified by a unique combination of surface structure, cover, and human activity at the local scale ($10^2$ to $10^4$ m). Originally designed for urban heat island research, LCZ classification also provides significant information, such as the pervious surface fraction and surface albedo values of each zone, in applications to green infrastructure, population assessment, and ecosystem processes \citep{stewart2012lczurban}.

{\bf Transfer Learning with Sliced Wasserstein Distance (SWD-transfer)}~~ \label{sec:swddescription}
Our work builds on a recent method that trains a Y-shaped neural network~\citep{rostami2019fewshot} to transfer knowledge from the EO to the SAR domain. The network consists of two convolutional neural networks with identical architectures as EO and SAR encoders. The outputs of the two encoders map to a shared embedding space that is then fed into a classification network. First, the network is trained to classify a large amount of EO data to obtain a discriminative embedding produced by the EO encoder.  
Then, using a small amount of labeled and a large amount of unlabeled SAR data, the network is trained to align the distribution produced by the EO encoder with that created by the SAR encoder. The minimization of discrepancy between the encoded distributions is achieved via minimization of the sliced Wasserstein distance. SWD computes an approximation of the distance between the two distributions by projecting them along a set of random vectors and computing the mean over the 1-dimensional Wasserstein distances of those random projections. More formally, SWD computes
\begin{equation}
    \vspace{-.5pt}
    \mathcal{D}^2(p_{\mathcal{S}}, p_{\mathcal{T}}) \approx \frac{1}{L} \sum_{l=1}^{L} \sum_{m=1}^{M} |\langle \gamma_l, \phi(\mathbf{x}_{s_l[i]}^\mathcal{S}) \rangle - \langle \gamma_l, \psi(\mathbf{x}_{t_l[i]}^\mathcal{T}) \rangle|^2 \enspace,
    \vspace{-.5pt}
\end{equation}
where $\mathcal{S}$ and $\mathcal{T}$ are the source and target domain respectively, $L$ is the number of random projections~$\gamma$, $\phi(\cdot)$ and $\psi(\cdot)$ are the  encoding functions, and $s_l[i]$ and $t_l[i]$ are sorting indices. The full approach (see \cite{rostami2019fewshot} for details) optimizes the objective
\begin{equation} \label{eq:optim}
\begin{split}
\vspace{-.5pt}
    \hspace{-6em}\min_{\mathbf{u, v, w}} & \frac{1}{N}\! \sum_{i=1}^N 
    \mathcal{L}(h_\mathbf{w}(\phi_{\mathbf{v}}(\mathbf{x}_i^s)), \mathbf{y}_i^s) + 
    \sum_{i=1}^O \mathcal{L}(h_\mathbf{w}(\psi_{\mathbf{u}}(\mathbf{x}_i^t)), \mathbf{y}_i^t) \\[-.5em]
    &+ \alpha \mathcal{D}^2(\phi_{\mathbf{v}}(p_\mathcal{S}(\mathbf{X}_{\mathcal{S}})), \psi_{\mathbf{u}}(p_{\mathcal{T}}(\mathbf{X}_{\mathcal{T}}))) +  \beta \sum_{j=1}^k \!\gamma \mathcal{D}^2(\phi_{\mathbf{v}}(p_\mathcal{S}(\mathbf{X}_{\mathcal{S}}|C_j)), \psi_{\mathbf{u}}(p_{\mathcal{T}}(\mathbf{X}_{\mathcal{T}})|C_j)) \enspace,\hspace{-4em}
    \vspace{-.5pt}
\end{split} 
\end{equation}

where the first two terms are the classification losses for each domain with $\mathcal{L}$ being the cross-entropy loss, the third term is the distribution alignment term over the unlabeled data, and the final term is a class conditioned alignment of distribution to ensure correct matching of class modes.

{\bf Normalizing Embedding Spaces}~~
In applying SWD-transfer to the Sentinel-1 and Sentinel-2 data, we found that it  is sensitive to the scale of the input data. We conjecture that not only is the scale of the input of significance, but so is the scale across {\em all} layers of the neural network. Normalization is a common technique in neural networks to ensure that layer outputs with diverse ranges will have a proportionate impact on the prediction. The original work suggests the use of batch normalization \citep{ioffe15batchnorm}, which computes mean and variance for every training batch and uses them for normalization. However, the wide range of the spectrum of the SAR data might distort estimates of those distribution measures. Instead, we consider that the specific nature of SAR data might benefit from using instance normalization~\citep{ulyanov16instance}, which computes a per-data-point mean and variance to prevent instance-specific mean and covariance shifts.

{\bf Supervised Contrastive Learning}~~ \label{sec:supcon}
We hypothesize that explicitly modeling the embedding space that is used for alignment can improve the performance of SWD-based transfer. To enforce a specific structure in the shared embedding space, we explore the impact of a recent supervised constrastive learning approach called SupCon~\citep{khosla2020scl}. Intuitively, contrastive learning enforces that data points from the same class are mapped closer together in embedding space and points from classes that are different ought to be pushed away from each other. By incorporating the contrastive loss into SWD-transfer, we hope to obtain a more discriminative embedding space and cleaner class-conditioned distribution boundaries in problems with large numbers of classes. Concretely, we add the following term to the optimization in Equation~\ref{eq:optim} when training the EO classifier:
\begin{equation}
\vspace{-.5pt}
    \mathcal{L}_{C} = \sum_{i \in I} \frac{1}{|P(i)|} \sum_{p \in P(i)} \log \frac{\exp(\mathbf{z}_i \times \mathbf{z}_p / \tau)}{\sum_{a \in A(i)} \exp(\mathbf{z}_i \times \mathbf{z}_a / \tau)} \enspace ,
    \vspace{-.5pt}
\end{equation}
where $\mathbf{z}_k = \phi_{\mathbf{v}}(\mathbf{x}_k)$, $i \in I$ is the index of an augmented data sample (e.g. rotation or translation of the original image) from a multiviewed batch, $A(i) \equiv I \setminus i$, and $P(i) \equiv \{p \in A(i)\!: \tilde{\mathbf{y}}_p = \tilde{\mathbf{y}}_i\}$ is the set of all positives in the batch distinct from $i$. A multiviewed batch is created by applying two random augmentations to every sample in a batch yielding a new batch of twice the original size.

\section{Experiments \& Results} 
\label{sec:experiments}

\begin{figure}[b!]
      % \captionsetup[subfigure]
      \centering
        \begin{subfigure}{1.0\linewidth}
        \centering
        \includegraphics[width=6.5cm, clip]{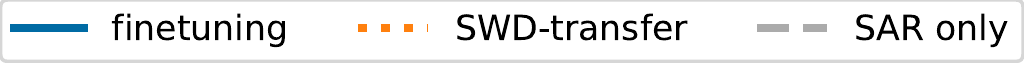}\\
      \end{subfigure} 
      \\
      \begin{subfigure}{0.32\textwidth}
        \centering
          \includegraphics[height=3.1cm,clip,trim=0in 0in 0in .4in]{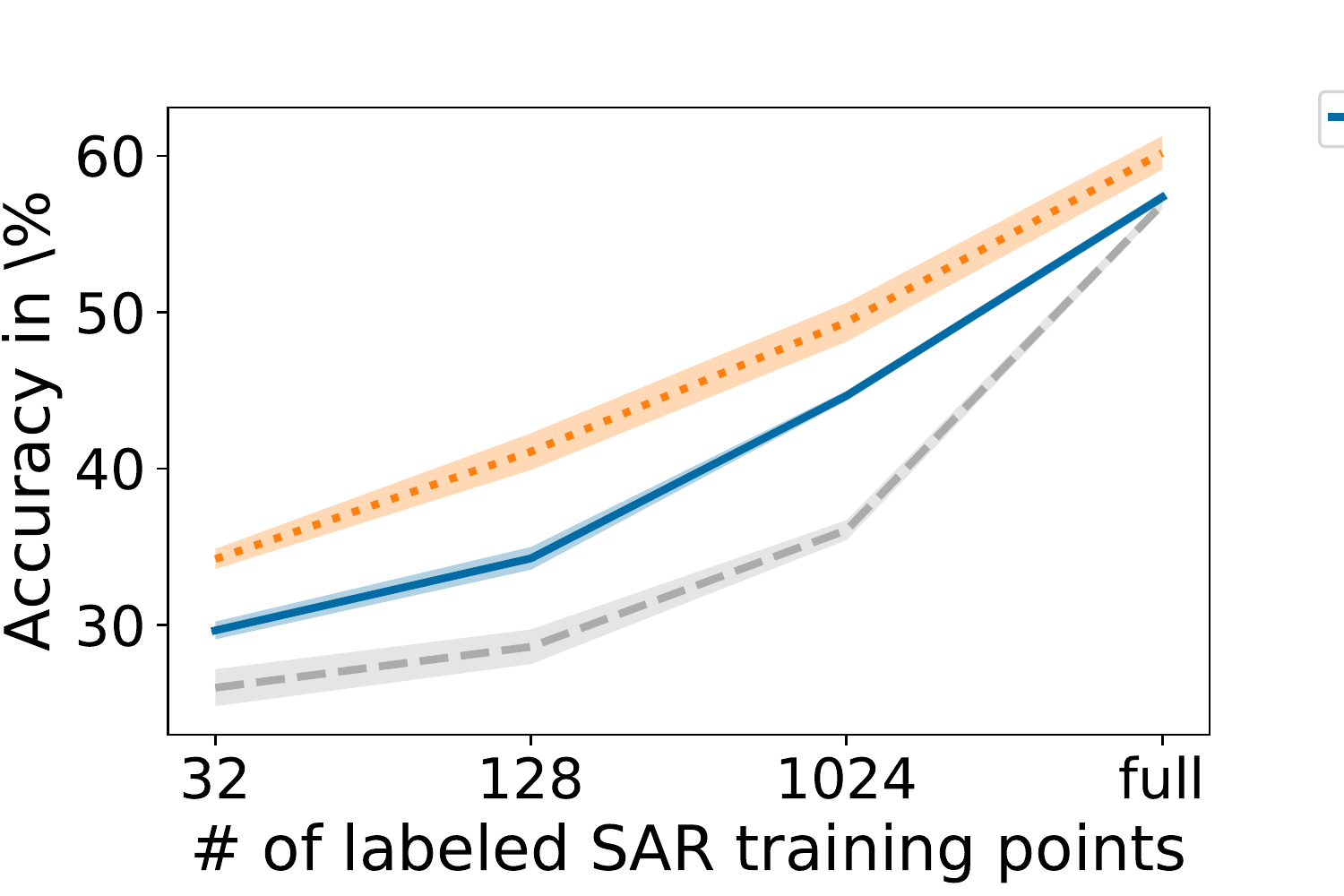}\\
          {(a) Low-rise Classes}
      \end{subfigure}
      \hfill
      \begin{subfigure}{0.32\textwidth}
        \centering
        \includegraphics[height=3.1cm, trim={0.7cm 0.0cm 0.0cm .4in}, clip]{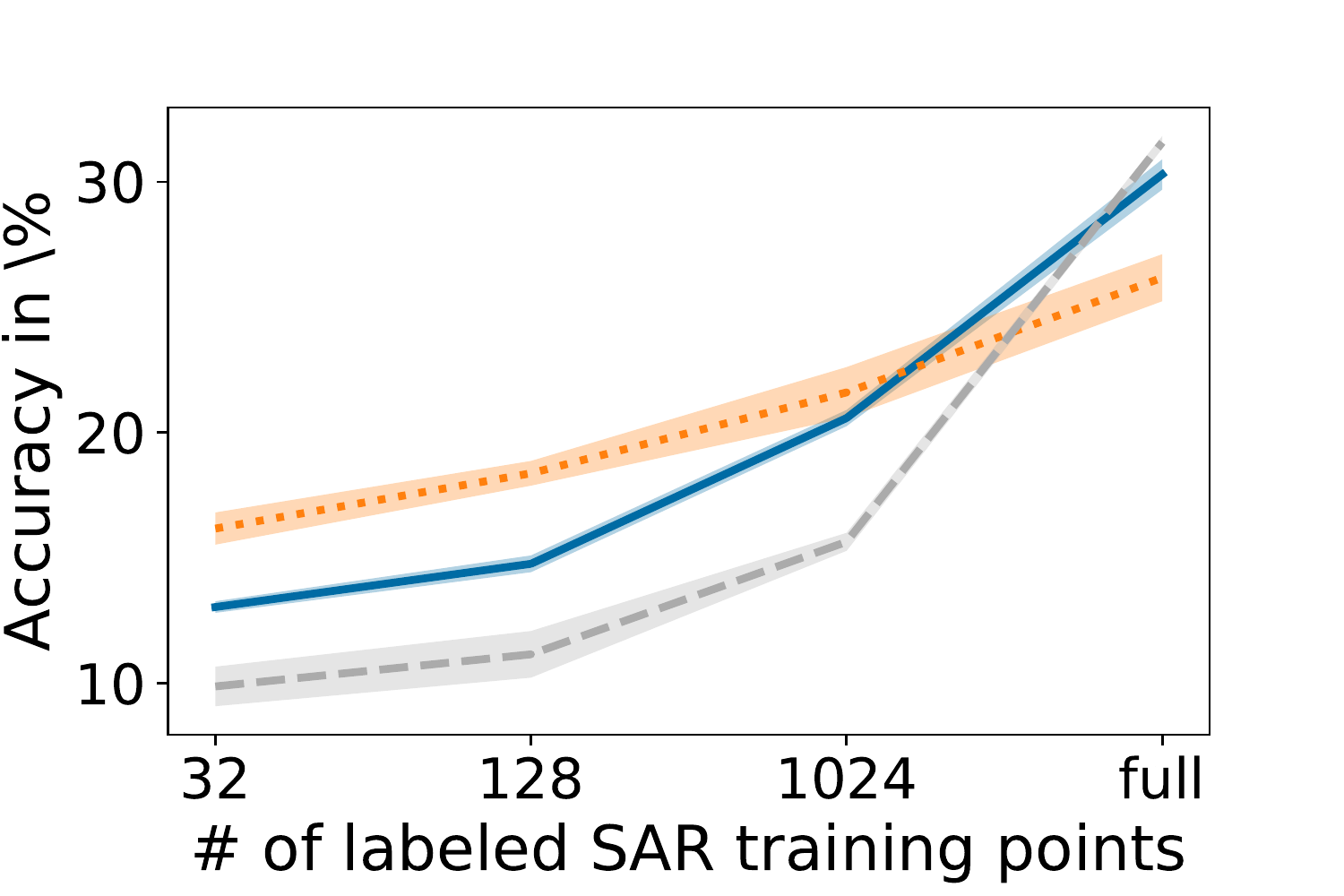}\\
        {(b) Urban Classes}
      \end{subfigure}
      \hfill
      \begin{subfigure}{0.32\textwidth}
        \centering
        \includegraphics[height=3.1cm, trim={0.7cm 0.0cm 0.0cm .4in}, clip]{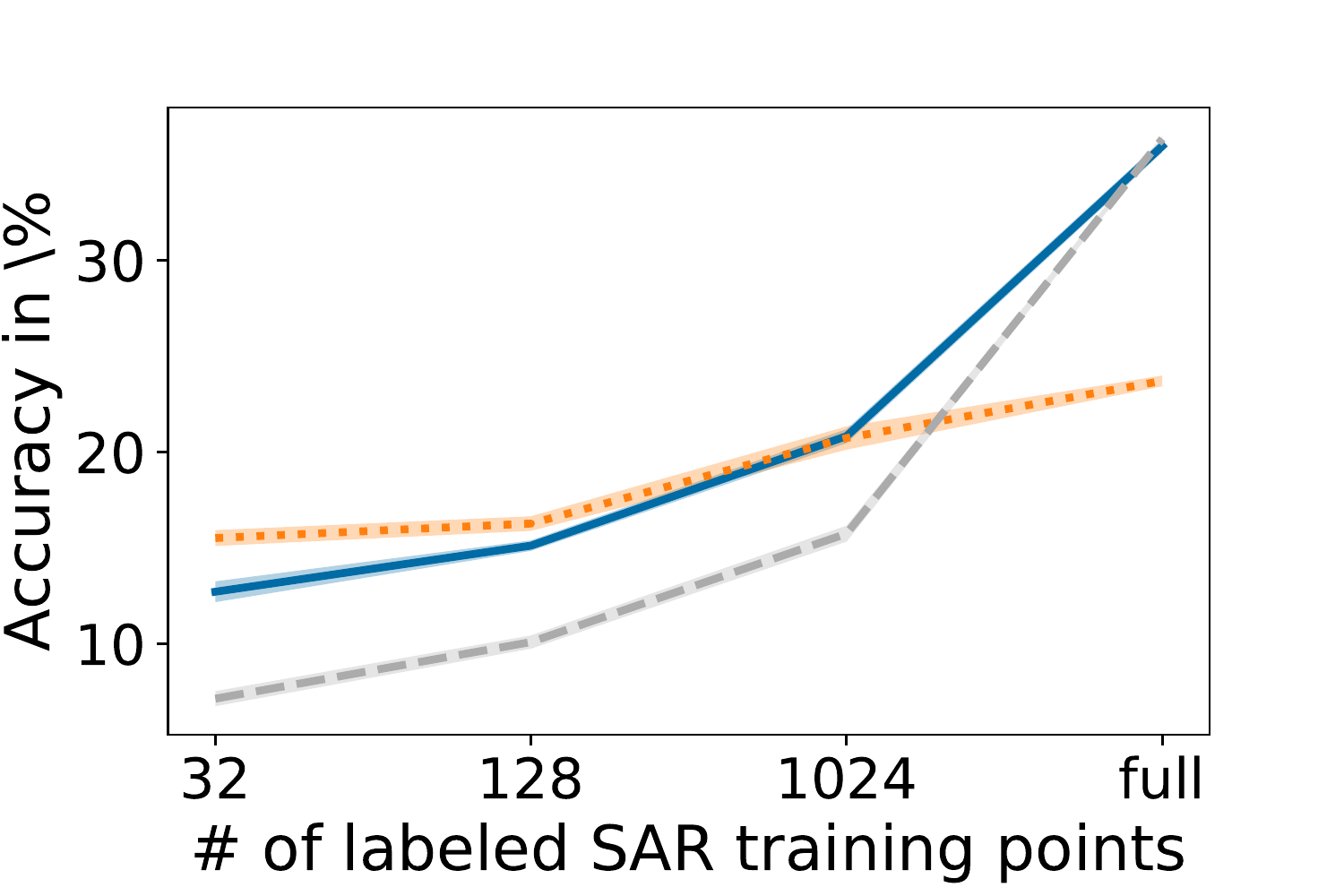}\\
        {(c) All Classes}
      \end{subfigure}
      \caption{Mean test accuracy and standard error for varying numbers of classes. SWD-transfer outperforms the baselines in the low data regime but is overtaken when the full dataset is available.}
      \label{fig:classCurves}
\end{figure}

We consider three configurations of the So2Sat dataset to show results on different levels of difficulties: four classes of different regions that contain \textit{low-rise} buildings, all $10$ \textit{urban} classes, and the full dataset with \textit{all} $17$ classes. For the EO images, we compute an RGB representation of the image, and for the SAR data, we use the real and imaginary parts of the unfiltered VV channel. Further, we compare three different methods: (a) training a single neural network on the EO data and then \textit{fine-tuning} it on the SAR domain, (b) training a classification network on the available labeled \textit{SAR data only}, (c) and \textit{SWD-transfer}~\citep{rostami2019fewshot}, as described in Section~\ref{sec:swddescription}. To evaluate effectiveness in the few-shot regime, we limit the number of available labeled SAR datapoints to $\textit{32}$, $\textit{128}$, $\textit{1024}$, or use the \textit{full} dataset. According to standard practice, we use train, validation, and test splits of the data, and report mean and standard error of the test accuracy over five random seeds. 

{\bf Scaling Sliced Wasserstein Transfer to Multi-Class Problems}~~ \label{sec:multiclass}
The first question that we want to answer is whether or not the SWD-transfer method scales to multi-class problems. For this, we consider the three dataset settings (low-rise, urban, all) to evaluate whether adding more classes to the problem decreases the relative performance compared to the baselines. The results are depicted in Figure~\ref{fig:classCurves}. First, we see that across all dataset settings, SWD-transfer outperforms the baselines in the low-data regime. With more classes and large amounts of data available, SAR-only training and fine-tuning outperform the SWD-transfer method. The results provide evidence that the SWD-transfer is indeed able to generalize to challenges with a large number of classes. However, the performance gap decreases as the number of classes increases, which is likely due to the learned EO embedding not being sufficiently discriminative.

\begin{figure} [t]
    \begin{minipage}[t]{.62\textwidth}
      % \captionsetup[subfigure]
      \centering
        \begin{subfigure}{1.0\textwidth}
        \centering
        \includegraphics[width=8cm]{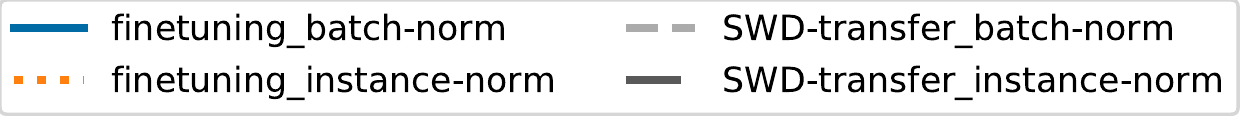}
      \end{subfigure} 
      \\
      \begin{subfigure}{0.49\textwidth}
        \centering
          \includegraphics[height=2.9cm,clip,trim=0in 0in .5in .4in]{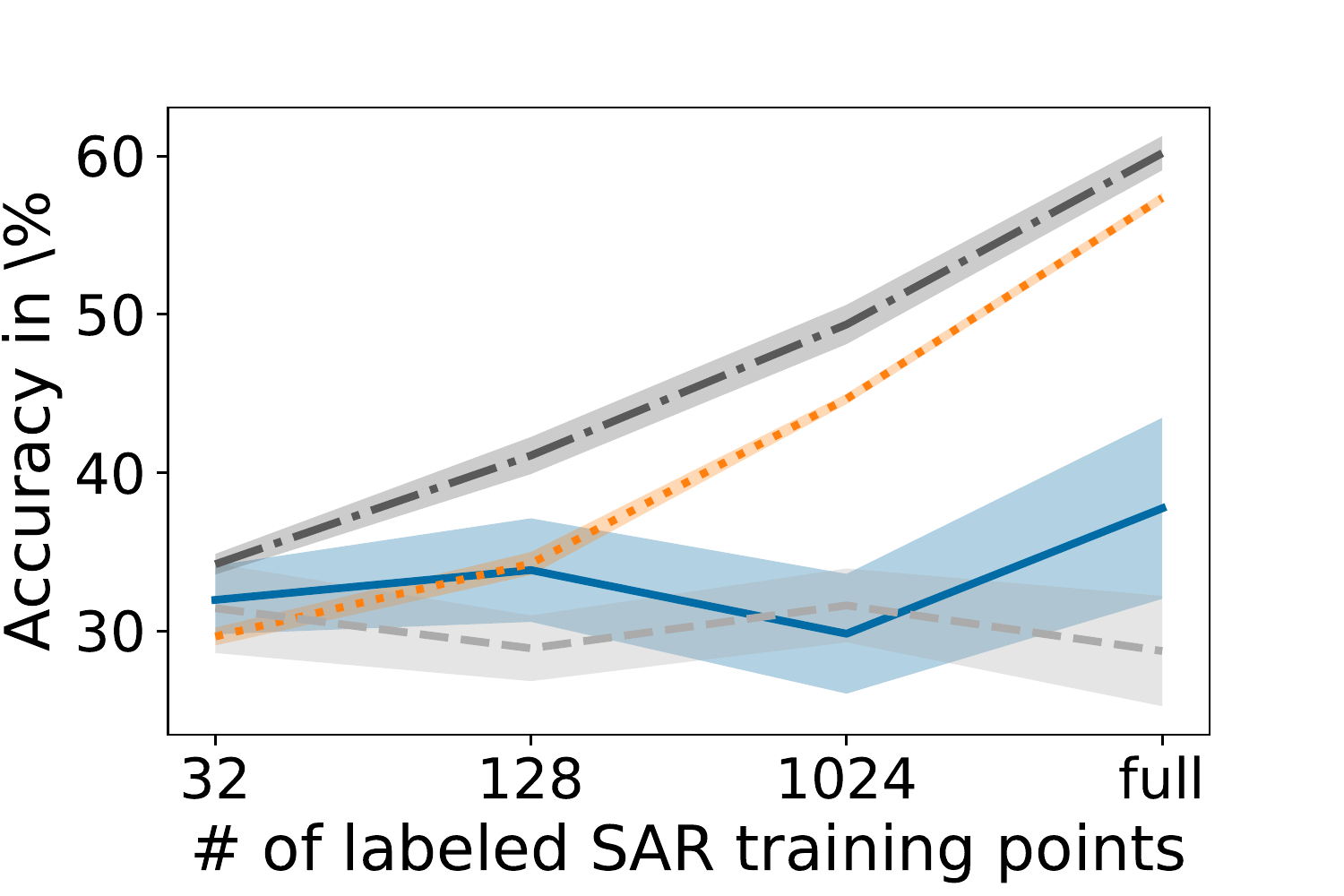}
          {(a) Low-rise Classes}
      \end{subfigure}
      \hfill
      \begin{subfigure}{0.49\textwidth}
        \centering
        \includegraphics[height=2.9cm, trim={0.7cm 0.0cm 0.0cm 0.4in}, clip]{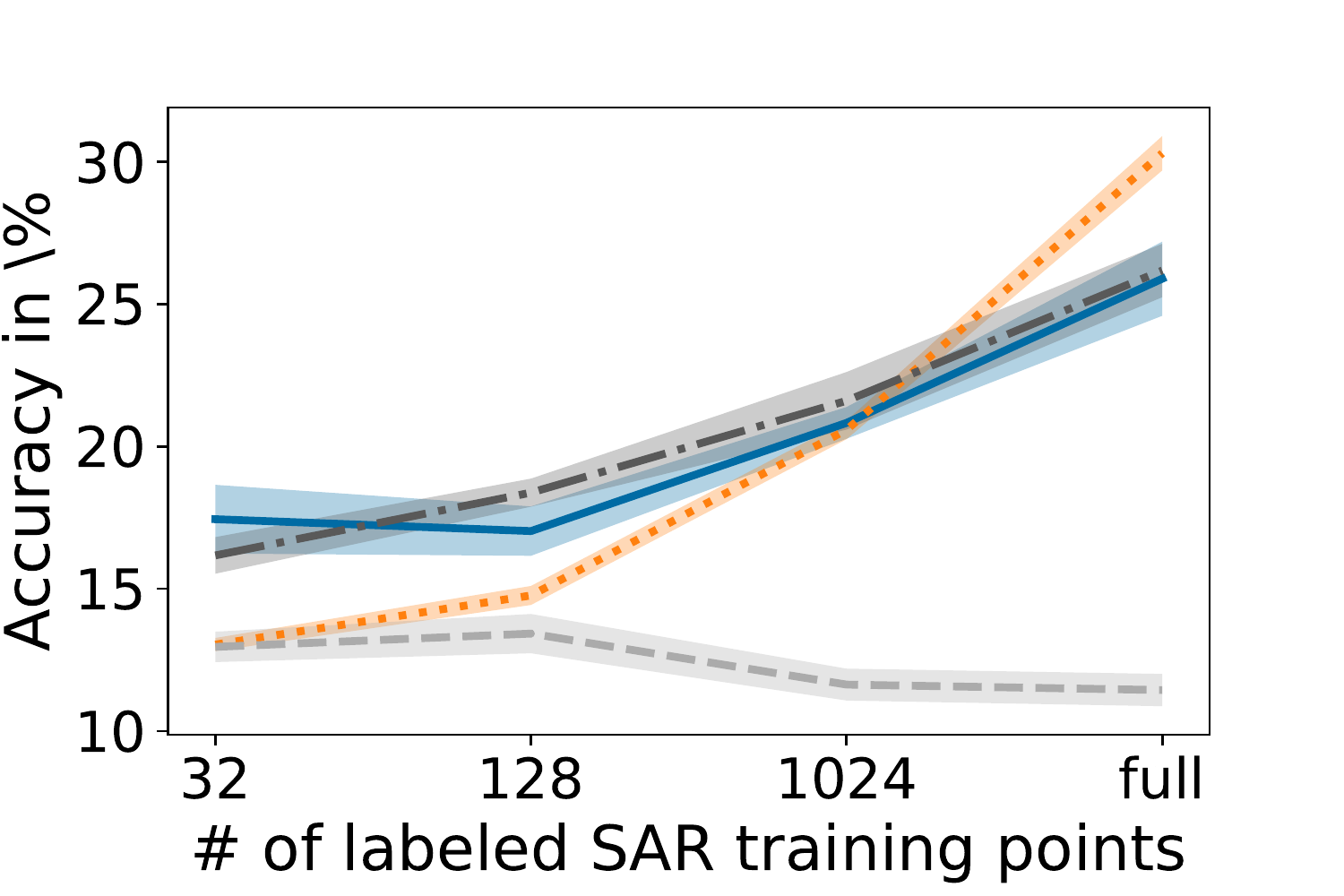}
        {(b) Urban Classes}
      \end{subfigure}
      \caption{Mean test accuracy and standard error of batch and instance normalization for fine-tuning and SWD transfer. Instance-normalization consistently stabilizes SWD-transfer training and leads to improved performance on both methods.}
      \label{fig:normalizationCurves}
    \end{minipage}%
    \hfill
    \begin{minipage}[t]{.34\textwidth}
      % \captionsetup[subfigure]
      \centering
        \begin{subfigure}{\textwidth}
        \centering
        \includegraphics[width=4.8cm, clip]{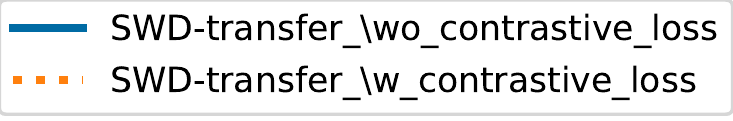}
      \end{subfigure} 
      \begin{subfigure}{\textwidth}
        \centering
          \includegraphics[height=2.9cm, clip,trim=0in 0in .5in .4in]{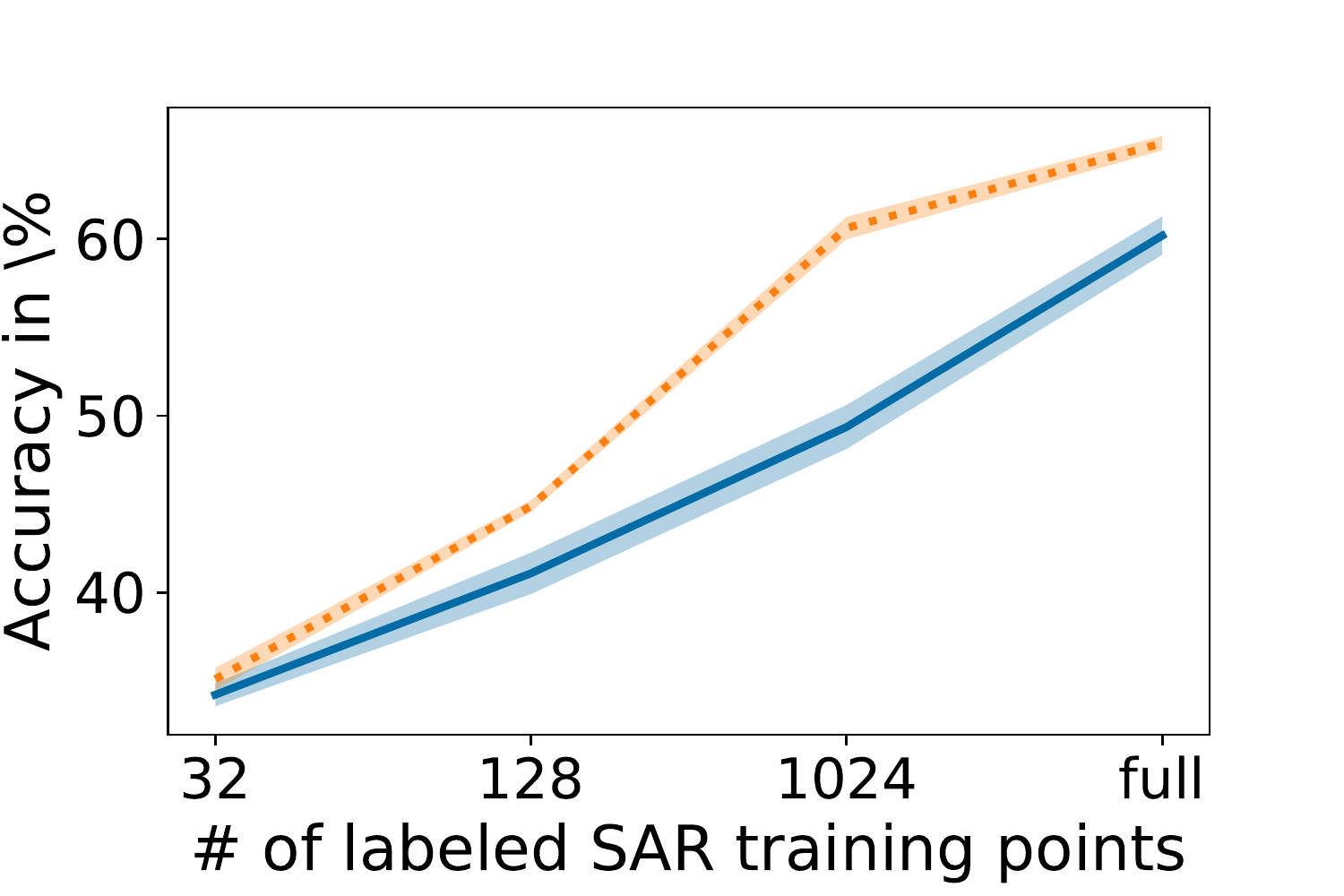}~\\
          {Low-rise Classes}
      \end{subfigure}
      \caption{Mean test accuracy and standard error when pretraining without or with contrastive loss. The latter increases performance.}
      \label{fig:contrastiveCurves}
    \end{minipage}
    %\vspace{-4pt}
\end{figure} 

{\bf The Importance of Normalization in Sliced Wasserstein Transfer}~~
This experiment explores how layer normalization can affect the performance of SWD-based transfer. Figure~\ref{fig:normalizationCurves} compares the performance of the SWD-transfer model and the fine-tuning approach when trained with batch or instance normalization. In most cases, SWD-transfer with batch normalization is not better than random guessing and instance normalization consistently stabilizes training and outperforms its counterpart. Additionally, instance normalization can improve the performance of the fine-tuning method. While in many cases the fine-tuning method works well with batch normalization, SWD-based transfer seems to \textit{require} instance normalization. This suggests that instance normalization is necessary in order for the approach to learn the embedding distribution alignment for SAR data.

{\bf Embedding Modeling during Pretraining using Contrastive Learning}~~
Finally, we analyze the effect of modeling the shared embedding space in order to explicitly obtain more discriminative embeddings. We applied the SupCon method (Section~\ref{sec:supcon}) to the EO training cycle, while the SAR training remained unchanged from the original SAR-transfer formulation. Results on the low-rise dataset setting (Figure~\ref{fig:contrastiveCurves}) shows that the SupCon-pretrained SWD-based transfer network outperforms the original SWD-transfer method across all data regimes. This presents evidence that shaping the internal representations of the neural network via contrastive learning facilitates the alignment of the distributions by creating more discriminative embeddings. 

\section{Conclusion and Future Work}

We identified several shortcomings of the original SAR-transfer method when applied to more realistic SAR data, and developed enhancements to compensate, moving towards EO-to-SAR transfer in more practical settings. 
The additions of instance normalization and supervised contrastive learning provide significant improvements on the base method, bringing us closer to deploying the approach in the wild. However, we highlight that none of the results achieve the very high accuracies required for automated land cover mapping, leaving much room for improvement. In the future, applying semi-supervised contrastive learning during the SAR training phase is an interesting idea to explore. 

%\pagebreak 

\section*{Acknowledgements}

We are grateful for the helpful technical discussions with Ryan Soldin and J.P.~Clark on this work.
The research presented in this paper was partially supported by Lockheed Martin Space, the Vagelos Integrated Program in Energy Research (VIPER) at Penn, the DARPA Lifelong Learning Machines program under grant FA8750-18-2-0117, the DARPA SAIL-ON program under contract HR001120C0040, the DARPA ShELL program under agreement HR00112190133, and the Army Research Office under MURI grant W911NF20-1-0080. Any opinions, findings, and conclusion or recommendations expressed in this material are those of the authors and do not necessarily reflect the view of Lockheed Martin, DARPA, the Army, or the US government.

\bibliographystyle{plainnat}
\bibliography{references}

\end{document}